\documentclass{llncs}
\usepackage{makeidx}  
\usepackage{graphicx}
\begin{document}
\mainmatter              
\title{High-resolution medical image synthesis using progressively grown generative adversarial networks}
\titlerunning{High-resolution medical image synthesis using PG-GANs}  
%
\author{Andrew Beers\inst{1} \and James Brown\inst{1} \and
Ken Chang\inst{1} \and J. Peter Campbell\inst{2} \and Susan Ostmo\inst{2} \and Michael F. Chiang\inst{2,3} \and
Jayashree Kalpathy-Cramer\inst{1,4}}
\authorrunning{Beers et al.} 
%
\tocauthor{Andrew Beers and James Brown and Ken Chang and J. Peter Campbell and Susan Ostmo and Michael F. Chiang and
Jayashree Kalpathy-Cramer}
\institute{Athinoula A. Martinos Center for Biomedical Imaging, Depart of Radiology, Massachusetts General Hospital,  Charlestown, MA, USA,\\
\and Department of Ophthalmology, Casey Eye Institute, \\Oregon Health \& Science University, Portland, OR
\and Department of Medical Informatics and Clinical Epidemiology, \\Oregon Health \& Science University, Portland, OR
\and Massachusetts General Hospital \& Brigham and Women’s Hospital\\ Center for Clinical Data Science, Boston, MA
}

\institute{}

\maketitle              

\begin{abstract}
Generative adversarial networks (GANs) are a class of unsupervised machine learning algorithms that can produce realistic images from randomly-sampled vectors in a multi-dimensional space. Until recently, it was not possible to generate realistic high-resolution images using GANs, which has limited their applicability to medical images that contain biomarkers only detectable at native resolution. Progressive growing of GANs is an approach wherein an image generator is trained to initially synthesize low resolution synthetic images (8x8 pixels), which are then fed to a discriminator that distinguishes these synthetic images from real downsampled images. Additional convolutional layers are then iteratively introduced to produce images at twice the previous resolution until the desired resolution is reached. In this work, we demonstrate that this approach can produce realistic medical images in two different domains; fundus photographs exhibiting vascular pathology associated with retinopathy of prematurity (ROP), and multi-modal magnetic resonance images of glioma. We also show that fine-grained details associated with pathology, such as retinal vessels or tumor heterogeneity, can be preserved and enhanced by including segmentation maps as additional channels. We envisage several applications of the approach, including image augmentation and unsupervised classification of pathology.
\keywords{deep learning, generative adversarial networks, unsupervised learning, image synthesis, retina, glioma}
\end{abstract}

\section{Introduction}
Synthesis of medical images has many applications. One application is image augmentation, whereby a small dataset with low diversity is amplified by approximating and randomly sampling the underlying data distribution. This can facilitate the training of more robust machine learning algorithms by dramatically increasing the size and heterogeneity of the training dataset. This is especially useful in scenarios where the patient population is small, such as for rare diseases \cite{chang2017institutionally}. Secondly, in multi-institutional studies, there may be differences in imaging protocols such that certain imaging modalities are missing for a subset or all of the patients. It would therefore be desirable to produce a complete set of data by synthesizing the missing modalities (or even additional modalities) for all patients to improve algorithm performance \cite{van2015does}. For example, brain tumor segmentation performs best when multiple MR modalities (such as T1, T2, and FLAIR) are used \cite{havaei2016hemis}. Lastly, certain imaging modalities have downsides to their acquisition. For example, CT and PET imaging impart a high dose of radiation to the patient. Effective synthesis of CT from other imaging modalities would avoid unnecessary radiation exposure \cite{nie2017medical} and reduce the costs incurred due to instrument time and contrast agents \cite{buck2010economic}.

The biomedical imaging community stands to gain from robust image synthesis methods. However, two key challenges exist with existing methods. Current techniques are unable to generate high resolution images, which are important for diseases with subtle pathological features. Secondly, image synthesis methods often neglect certain biological features that are critical for diagnosis, and therefore need to be well-represented. Here, we propose a method of addressing both of these challenges. Generative adversarial networks (GANs) have been used to generate synthetic images of unprecedented realism and diversity \cite{goodfellow2014generative}. Applications in imaging - including biomedical imaging - have flourished, but have been confined to relatively small image sizes \cite{marchesi2017megapixel}. Recently, Karras \textit{et al.} devised a training scheme for GANs called progressive growing of GANs (PGGANs) that can create photorealistic images at high resolutions, with images up to 1024 $\times$ 1024 pixels being showcased in their work \cite{karras2017progressive}. However, their application was limited to common image benchmarking datasets such as celebrity faces and natural scenes. Application of PGGANs to biomedical data with clinically-relevant imaging biomarkers has yet to be explored.

In this paper, we propose an application of the PGGAN method to two classes of medical image: retinal fundus photographs with retinopathy of prematurity (ROP), and two-dimensional magnetic resonance images taken from a publicly-available, multi-modality glioma dataset (BraTS). We show that application of PGGANs to these data produces images at high resolution that are both realistic and phenotypically diverse. We also demonstrate that preservation of fine-grained details of pathology may be improved by having the generator produce a segmentation of a structure of interest, which is criticized alongside the raw image by the discriminator. Lastly, we show that the latent space encodes clinically relevant information by inverting the generator and transforming real images into low dimensional representations.

\section{Datasets}

\subsection{Retinopathy of Prematurity Dataset}
We used a dataset of 5,550 posterior pole retinal photographs collected as part of the ongoing multicenter Imaging and Informatics in Retinopathy of Prematurity (i-ROP) cohort study. All images were resized to 512 $\times$ 512 pixels. Segmentations of the retinal vasculature were produced by first converting images to grayscale and applying gamma adjustment and contrast-limited adaptive histogram equalization (CLAHE). A U-net \cite{ronneberger2015u} pre-trained on 200 manually segmented vessel images was used to perform segmentation of the vasculature, which operates on 48 $\times$ 48 pixel patches with an overlap of 8 pixels in the $x$ and $y$ directions.

\subsection{Brain Tumor Magnetic Resonance Imaging Dataset}

We utilized the Brain Tumor Segmentation (BraTS) Challenge 2017 training dataset, which contains magnetic resonance (MR) images from 75 low-grade and 210 high-grade glioma patients from 19 institutions, with varying scanner models, magnetic field strengths, echo times, and repetition times \cite{menze2015multimodal} \cite{bakas2017advancing}. The modalities included T1, T1 post-gadolinium, T2, and T2 FLAIR weighted MR imaging. All imaging was co-registered with respect to each patient, brain extracted, and isotropically resampled. 3 segmentation labels from 1-4 expert raters were provided:  1) edema, 2) non-enhancing/necrosis, and 3) contrast enhancing tumor. We preprocessed each MR volume to have zero mean and unit variance.

\section{Network Architectures}
\begin{figure}
\centering
\includegraphics[width=1.0\linewidth]{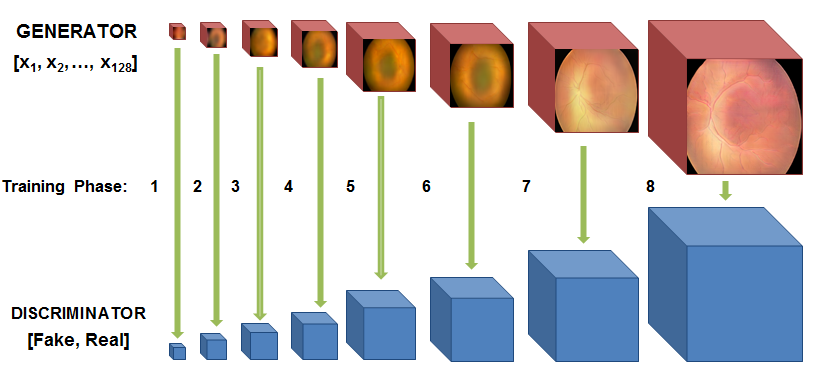}
\caption{A schematic for a progressively growing GAN (PGGAN). The GAN is trained in phases, with each phase adding an upsampling layer and a pair of convolutional layers to the both the discriminator and generator. Outputs from a single latent vector at each stage of the training are shown superimposed on the generator arm.}
\label{network}
\end{figure}
The PGGAN architecture trains a single GAN in a step-wise fashion (Figure \ref{network}). In our case, we initially train a GAN to produce realistic downsampled images at 4x4 pixel resolution from a 128-vector latent space via a pair of convolutional layers, and then discriminate between real and synthesize images via a symmetric pair of convolutional layers in the discriminator. We then iteratively add pairs of convolutional layers to both sides of the GAN architecture, that synthesize images at 8, 16, 32, 64, 128, 256, and 512 square pixels respectively. We follow the scheme suggested by Karras et al. for adding additional layers. When transitioning from generating lower to higher resolution images, we interpolate between nearest neighbor upsampling alone, and nearest neighbor upsampling followed by the newly-added pair of convolutional layers. Convolutional layers below 256 pixel resolution are set at 128 filters, while layers at and above 256 are set to 64 and 16 layers respectively. Training stages at resolutions below 512 pixels are trained with a 16 image batch size, while resolutions equal to and above 512 pixels are trained with an 8 image batch size. Both of these choices specific to higher resolutions were made in deference to GPU memory constraints. Models were trained using an Adam optimizer for 40,000 batches per resolution, with 20,000 iterations spent on the interpolation phase, and 20,000 iterations training without interpolation. We evaluated the loss for the GAN using a Wasserstein loss, and trained the generator and discriminator in equal proportion. For the retinopathy of prematurity dataset, we trained up to 512 pixel resolution, while for the glioma dataset, we trained up to 256 pixel resolution. All images were trained on an NVIDIA P100 GPU.

We also trained a network for encoding images using the same network architecture as the discriminator, with the exception of terminating in a 128-vector output with a mean-squared error cost function. This network was fed synthetic images generated from the already-trained PGGAN, and trained to predict the latent vector which generated that synthetic image. It was trained in a non-progressive manner for 20,000 iterations under the same training conditions as the primary PGGAN.

\section{Results}

\begin{figure}
\centering
\includegraphics[width=1\linewidth]{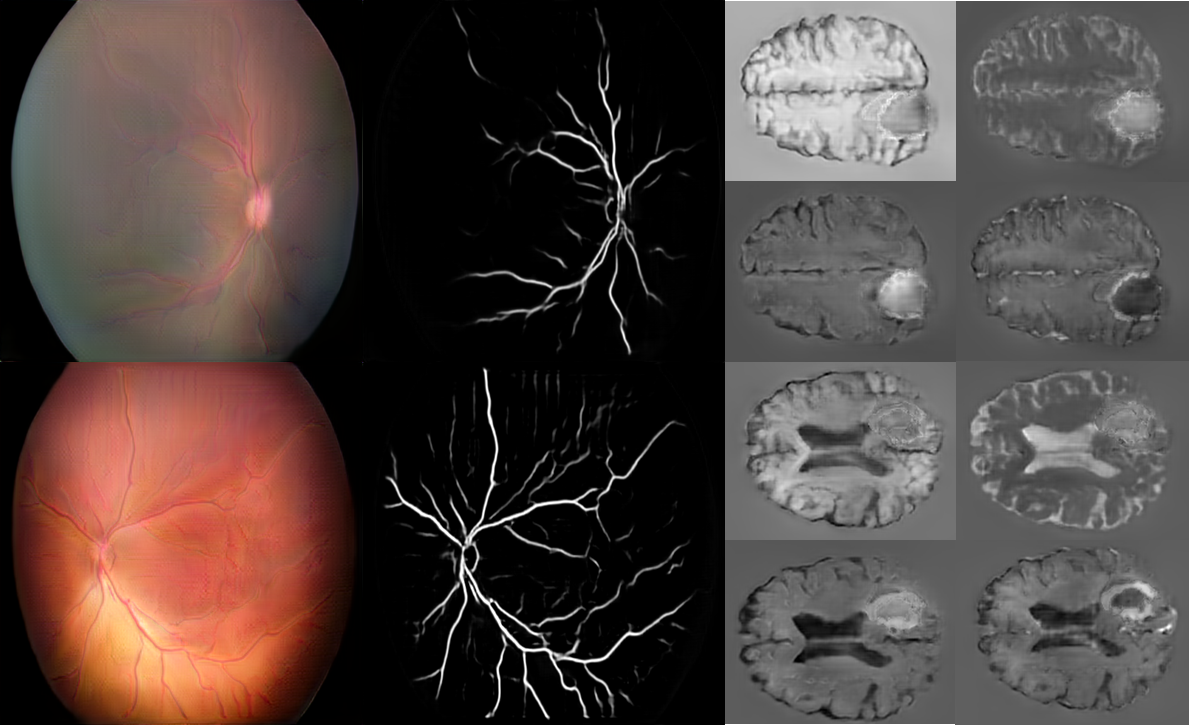}
\caption{Images generated by a PGGAN trained on a retinal fundus dataset (left) and an by the PGGAN trained on multi-modal glioma MRI data (right). Retinal images show synthetic fundus images and synthetic vessel maps, while MRI images show synthetic 2D MRI slices (modalities clockwise from top-right: T2, post-contrast T1, FLAIR, and pre-contrast T1).}
\label{high_res_images}
\end{figure}

\subsection{Image Quality}

In both retinal fundus images and glioma multi-modality MRI images, images of high quality and variation were produced by the PGGAN architecture (Figure \ref{high_res_images})	. Qualitatively, some unrealistic images were produced, moreso in the MRI slice case. Both examples suffered from the appearance of overly-distinct edges in pathological tissue, likely caused by the requirement for the generator to create segmentation maps in addition to anatomy. Upon close inspection, minor checker-boarding artifacts could be found in the retinal images.

In the case of the retinal fundus images, we further tested whether data generated by the PGGAN would be interpreted similarly by algorithms trained only on real retinal data. Specifically, we took a state of the art retinal vessel segmentation algorithm, applied it to synthetic fundus images generated by the GAN, and evaluated the differences between the algorithm's computed segmentations and the segmentation map generated automatically by the GAN. We found that segmentation maps generated from from the GAN could accurately predict the segmentation maps generated from the segmentation algorithm with the AUC of 0.97, suggesting that the vessels and segmentation maps generated by the GAN are in the modeling context indistinguishable from vessels in real images.

\subsection{Effect of Segmentation Channels}

\begin{figure}
\centering
\includegraphics[width=1.0\linewidth]{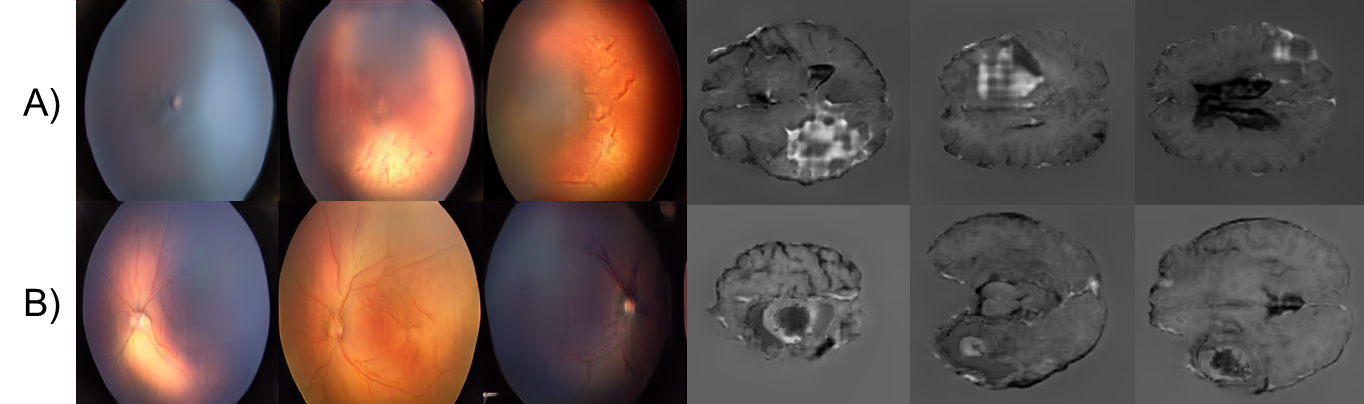}
\caption{Both rows show synthetic images generated by PGGANs. Images in Row A show images generated from a PGGAN that did not train on vessel segmentation data or glioma segmentation data, while images on Row B were generated from a PGGAN that trained on segmentation data.}
\label{channels_compare}
\end{figure}

To accurately synthesize those areas of the image important to medical diagnosis, we included segmentation maps previously described as additional channels to the network. As shown in Figure \ref{channels_compare}, images trained without these segmentation maps often obscured details important to pathological diagnosis, such as global retinal structures in the fundus images, and biologically-plausible tumor structures in the MRI slices. 

\subsection{Image Variety}

\begin{figure}
\centering
\includegraphics[width=1\linewidth]{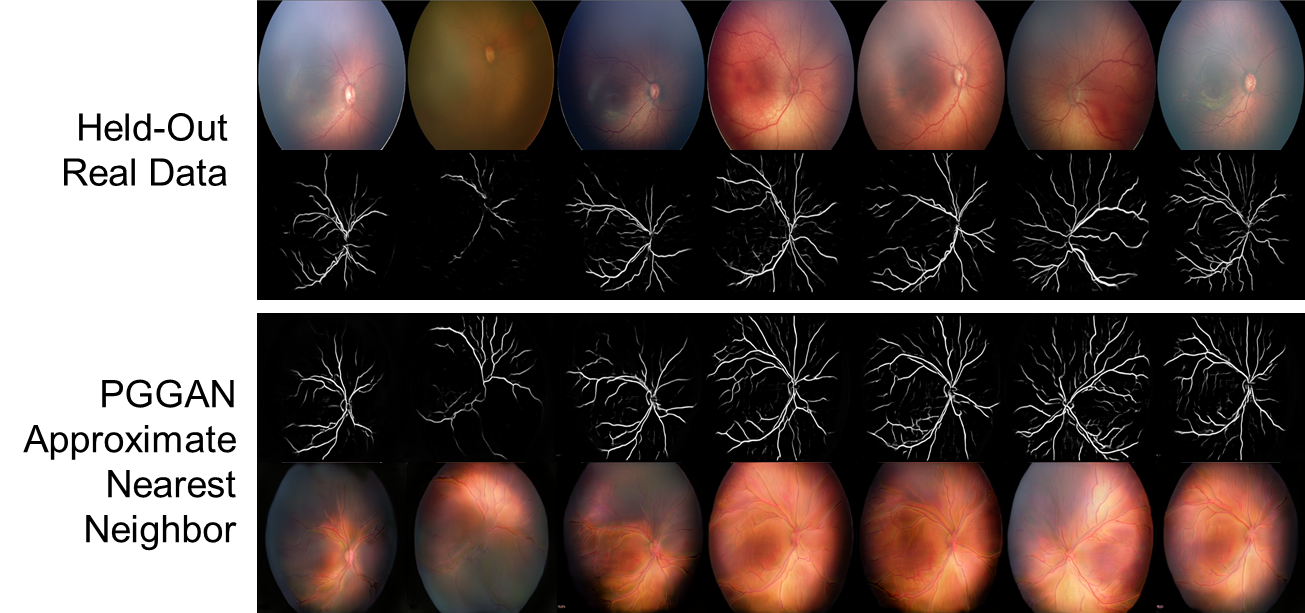}
\caption{The top rows consist of retinal fundus data held out from the training of the PGGAN, and their segmentations produced via a state of the art segmentation algorithm. The bottom row consists of images generated from latent vectors that were derived from the top row by an encoding neural network. These images are, approximately, the nearest neighbor in latent space to the true images in the top row.}
\label{decode_compare}
\end{figure}

Qualitatively, a variety of lighting conditions, vessel configurations, and optic disc locations could be seen generated in the retinal fundus images. In MRI images, a similar variety in slice location, tumor location, presence and absence of tumor, and distribution of necrosis, edema, and enhancing tumor could be found.

To further explore the variability of the synthetic images produced, we separately trained a neural network to determine the latent vector the produced a given input image, and used it to generate approximate nearest neighbors in the latent space of a test set of 200 real images held-out from the PGGAN training set. As shown in Figure \ref{decode_compare}, the neural network was able to qualitatively approximate the global vessel structure of each real image in the held-out set, suggesting that this PGGAN has the ability to generate vessel trees from outside its original training set.

\begin{figure}
\centering
\includegraphics[width=.8\linewidth]{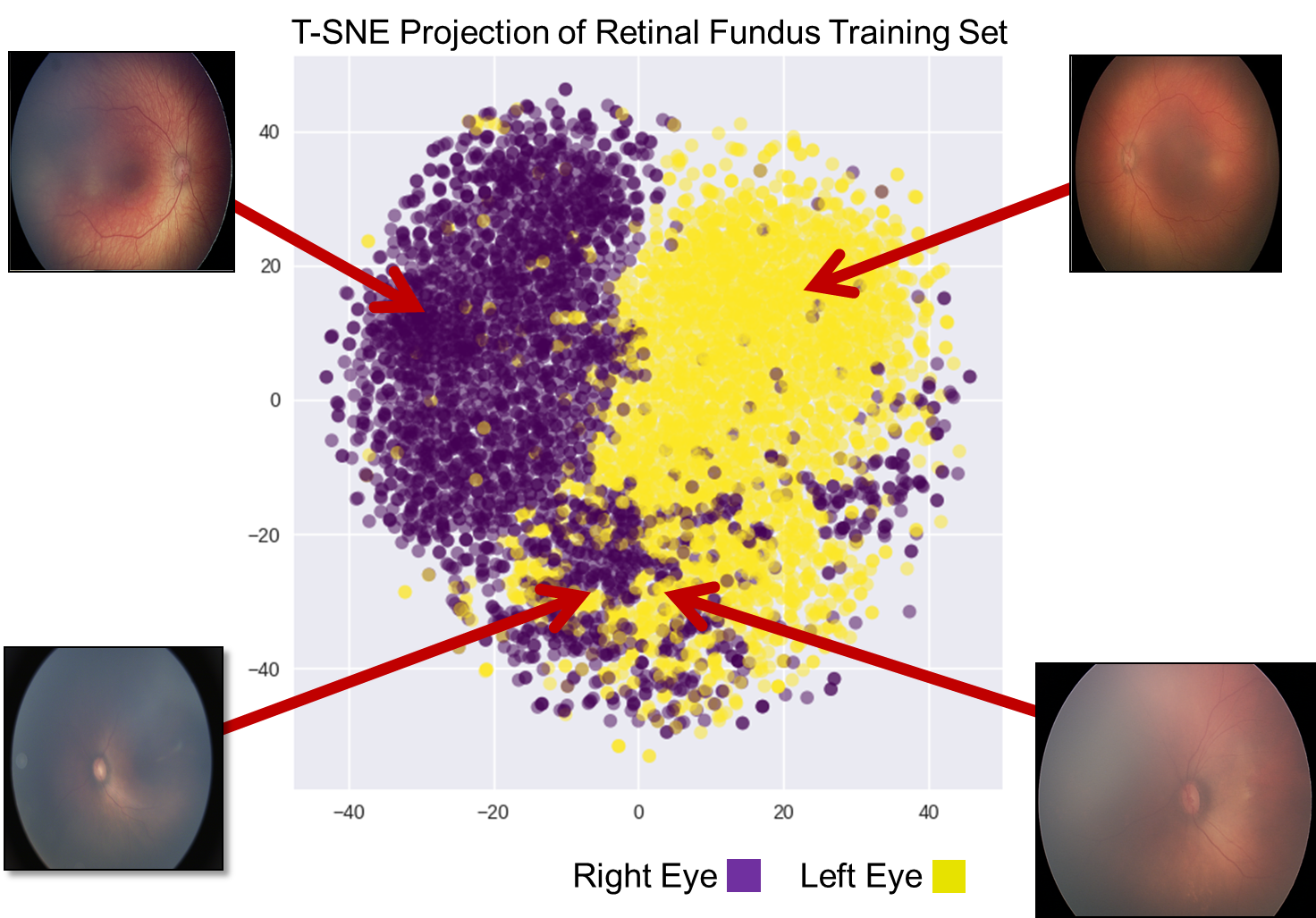}
\caption{Each point above represents an image from the training set for the retinal fundus PGGAN. These real images have been encoded into 128-dimensional latent vectors in the space of the PGGAN via a separate neural network, and those latent vectors have been projected onto a two-dimensional surface via t-Distributed Stochastic Neighbor Embedding (T-SNE). Each point has been color-coded according to whether its image represents a right eye or left eye.}
\label{tsne_retina}
\end{figure}

\subsection{Latent Space Expression}

We refer here to the latent space as the space of images generated by the 128-dimensional latent vector inputs to the PGGAN. Previous work has shown that the latent space of GANs often encodes semantic information about the images produced, and that latent vectors similar to each other in  latent space produce qualitatively similar output images \cite{donahue2016adversarial}. This result can be useful if one wishes to generate images of a certain phenotype, as they only need determine which subset(s) of latent space correspond to images generated with that phenotype, and sample accordingly.

To show that such qualities hold for PGGANs in medical imaging data, we trained a neural network to generate approximate latent vectors for each of the images in our training set. We then used t-Distributed Stochastic Neighbor Embedding (T-SNE) to create a two-dimensional reduced representation of the 128-dimensional space that these latent vectors occupied, and color-coded the results by simple morphological characteristics of the training data. One can see that a right eye / left eye distinction is primary in the latent space of the PGGAN generating retinal images in Figure \ref{tsne_retina}. Similarly, one can traverse through the axial slices of the brain by interpolating through images produced in the T-SNE projection of encoded slices from the BRATS training dataset (Supplement). Upon further exploration of the latent space, and additional annotation of input data, other associations may be found.

\section{Conclusion}
We present an application of the PGGAN method to retinal fundus and MRI data, specifically for the diseases of retinopathy of prematurity and glioma. This method can generate synthetic images of unprecedented size, and be used via its latent to space to learn imaging features in an unsupervised manner. Its application will open new avenues for synthetic image generation in medical imaging, which has thus far been limited by an inability to synthesize images at native resolution.

%
%

\bibliographystyle{splncs}
\bibliography{thebib}




\end{document}